\documentclass[letterpaper, 10 pt, conference]{ieeeconf}  % Comment this line out if you need a4paper

\IEEEoverridecommandlockouts                              % This command is only needed if 
\usepackage{algorithm}
\usepackage{algorithmic}
\usepackage[algo2e]{algorithm2e}
\usepackage{cite}
\usepackage{amsmath}
\usepackage{graphicx} 
\usepackage{lipsum}    
\usepackage{float}     
\usepackage{caption}   
\usepackage{booktabs}
\usepackage{amssymb}
\title{\LARGE \bf
    FRMD: Fast Robot Motion Diffusion with Consistency-Distilled Movement Primitives for Smooth Action Generation
}

\author{Xirui Shi$^{1,\dag}$ and Jun Jin$^{1,2}$% <-this % stops a space
\thanks{$^{1}$Department of Electrical and Computer Engineering, University of Alberta, Edmonton, Canada
 }%
\thanks{$^{2}$Alberta Machine Intelligence Institute (Amii), Edmonton, Canada
        }%
\thanks{$^{\dag}$Corresponding author:{\tt\small \ dalen.shi@ualberta.ca}}
}

\begin{document}

\maketitle
\thispagestyle{empty}
\pagestyle{empty}

%%%%%%%%%%%%%%%%%%%%%%%%%%%%%%%%%%%%%%%%%%%%%%%%%%%%%%%%%%%%%%%%%%%%%%%%%%%%%%%%
\begin{abstract}
We consider the problem of using diffusion models to generate fast, smooth, and temporally consistent robot motions. Although diffusion models have demonstrated superior performance in robot learning due to their task scalability and multi-modal flexibility, they suffer from two fundamental limitations: (1) they often produce non-smooth, jerky motions due to their inability to capture temporally consistent movement dynamics, and (2) their iterative sampling process incurs prohibitive latency for many robotic tasks. Inspired by classic robot motion generation methods such as DMPs and ProMPs, which capture temporally and spatially consistent dynamic of trajectories using low-dimensional vectors --- and by recent advances in diffusion-based image generation that use consistency models with probability flow ODEs to accelerate the denoising process, we propose \textbf{Fast Robot Motion Diffusion (FRMD)}. FRMD uniquely integrates Movement Primitives (MPs) with Consistency Models to enable efficient, single-step trajectory generation. By leveraging probabilistic flow ODEs and consistency distillation, our method models trajectory distributions while learning a compact, time-continuous motion representation within an encoder-decoder architecture. This unified approach eliminates the slow, multi-step denoising process of conventional diffusion models, enabling efficient one-step inference and smooth robot motion generation. We extensively evaluated our FRMD on the well-recognized Meta-World and ManiSkills Benchmarks, ranging from simple to more complex manipulation tasks, comparing its performance against state-of-the-art baselines. Our results show that FRMD generates significantly faster, smoother trajectories while achieving higher success rates.

\end{abstract}

%%%%%%%%%%%%%%%%%%%%%%%%%%%%%%%%%%%%%%%%%%%%%%%%%%%%%%%%%%%%%%%%%%%%%%%%%%%%%%%%
\section{Introduction}
Recently, diffusion models~\cite{chi2023diffusion} have gained increasing attention in robot learning due to their scalability~\cite{dasari2024ingredients} to complex tasks and their flexibility~\cite{dong2023aligndiff} in incorporating high-dimensional, multi-modal observations. Modern advancements~\cite{dasari2024ingredients, black2024pi_0} in embodied artificial intelligence have shown that an “action expert”~\cite{black2024pi_0} using diffusion models to generate robot actions can handle various manipulation tasks and that its multi-task capabilities scale up with dataset and model size. This highlights the promise of a unified model architecture that employs diffusion models as the action decoder (motion generation) for general-purpose robotic task solvers. However, existing diffusion-based robot motion generation methods still face two fundamental challenges. First, they often produce non-smooth, jerky motions because they fail to capture temporally consistent movement dynamics~\cite{carvalho2024motion}. Specifically, this limitation arises from the fact that these methods commonly generate raw action sequences (i.e., waypoints) without accounting for the trajectory-level temporal consistency imposed by the robot's structured dynamic constraints. Second, the iterative sampling process inherent to these diffusion models' denoising sampling process (Denoising Diffusion Probabilistic Model (DDPM)~\cite{ho2020denoising}, Denoising Diffusion Implicit Models (DDIM)~\cite{song2020denoising}) introduces significant latency when generating robot actions.

Inspired by classic dynamic-system–based robot motion generation methods, such as dynamic movement primitives (DMPs~\cite{li2023prodmp}) and probabilistic movement primitives (ProMPs~\cite{paraschos2013probabilistic}), which capture temporally and spatially consistent dynamic of trajectories using low-dimensional vectors, we propose rethinking diffusion-based robot motion generation by shifting from modeling the conditional action distribution at the raw action (waypoint) level to the trajectory level (movement primitives). This approach essentially learns the prior distribution in the trajectory parameter space. Moreover, instead of simply combing diffusion models with movement primitives (MPs) like previous methods~\cite{carvalho2024motion,scheikl2024movement}, to further accelerate the action denoising process, we incorporate recent advances in diffusion-based image generation~\cite{song2023consistency} that employ consistency models with Probability Flow ODEs~\cite{song2023consistency} and consistency distillation~\cite{song2023improved} for fast motion generation. These two effective ingredients enable our method to generate fast and smooth robot motions.

Specifically, we introduce Fast Robot Motion Diffusion (FRMD), a novel consistency-distilled movement primitives framework that integrates Consistency Models \cite{song2023consistency} and Probabilistic Dynamic Movement Primitives (ProDMPs) \cite{li2023prodmp} to achieve both high inference efficiency and smooth motion generation. Instead of directly generating raw actions via iterative denoising, we employ a movement primitives framework \cite{schaal2005learning, saveriano2023dynamic} to produce smooth motions with guaranteed initial conditions for planning consecutive action sequences. In this setup, the diffusion model predicts ProDMPs weight vectors to ensure structured and smooth trajectories. 

Moreover, unlike recent methods\cite{carvalho2024motion, scheikl2024movement} that simply combine diffusion models with movement primitives—which incur inference latency from multi-step denoising and limit their applicability in many robotic tasks—we observe that shifting from predicting raw actions (waypoints) to predicting trajectory parameters (movement primitives) naturally enables the use of consistency models\cite{song2023consistency} for accelerated inference. We employ consistency distillation \cite{song2023consistency,song2023improved} to train a model that directly maps noisy actions to movement primitive parameters. This single-step inference approach eliminates multi-step denoising while preserving the expressive power of Movement Primitive Diffusion (MPD) \cite{scheikl2024movement}, our teacher model. As a result, FRMD produces structured motion faster and smoother than traditional diffusion policies, broadening its applicability in many robotic tasks. Our contributions can be summarized as follows:
\begin{itemize}
\item We propose FRMD, a novel framework that combines Consistency Models with Movement Primitives, enabling fast inference while maintaining structured and smooth motion generation.
\item We introduce a consistency-distillation strategy to eliminate iterative denoising in diffusion-based motion generation, achieving one-step inference without sacrificing motion quality.
We extensively evaluate FRMD across multiple robotic manipulation tasks, demonstrating significant improvements in both inference speed and task success rates over existing diffusion-based methods.
\item By bridging structured motion representation and fast generative modeling, FRMD paves the way for real-time, high-quality robotic action generation, unlocking new possibilities in robot learning.
\end{itemize}

We evaluate FRMD on a diverse set of robotic manipulation tasks, ranging from simple to complex, using the MetaWorld \cite{yu2020meta} and ManiSkill \cite{mu2021maniskill} benchmarks following~\cite{lu2024manicm}. We compare FRMD with state-of-the-art diffusion policies, including the vanilla Diffusion Policy (DP) ~\cite{chi2023diffusion} and a recent method that combines diffusion models with movement primitives (MPD) ~\cite{scheikl2024movement}. The results show that FRMD achieves the highest success rate (64.8\%) while operating 10× faster than MPD (the teacher model) and 7× faster than DP, enabling the generation of fast and smooth robot motions.

\begin{figure*}[htbp]
    \centering
    \includegraphics[width=\textwidth]{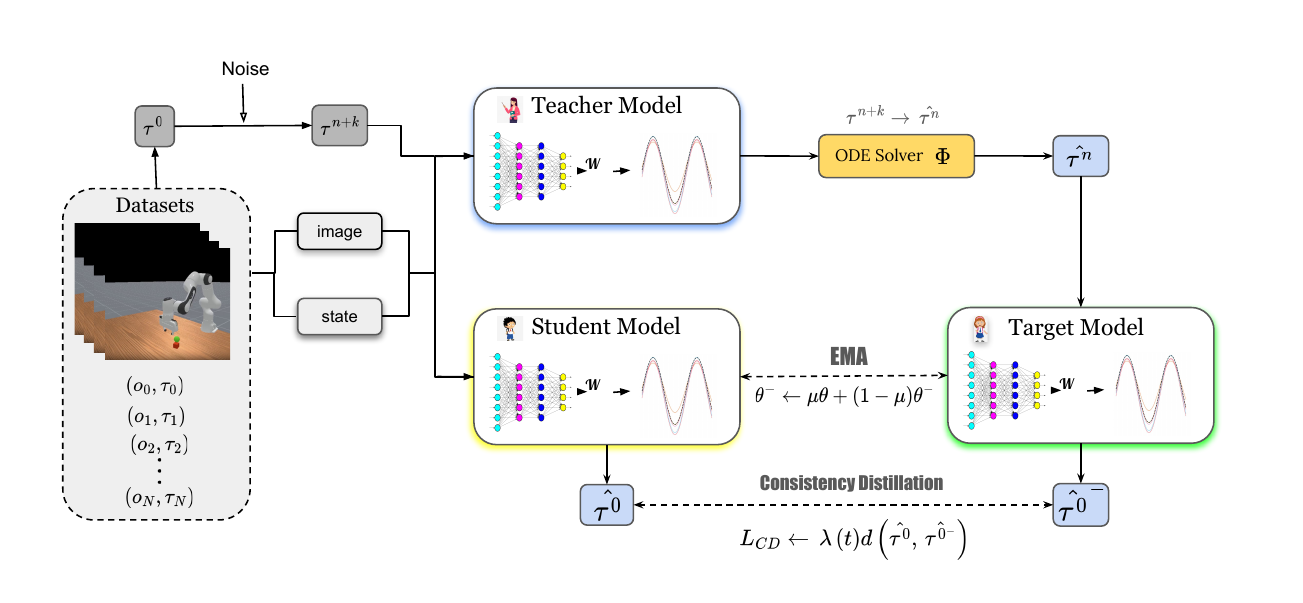}
    \caption{\textbf{Overview of FRMD Training Framework.} Given observations $o_i$, raw action sequence $\tau^0$ and initial state $(y_0, \dot{y_0})$ from the robot datasets, we first perform a forward diffusion to introduce noise over $n+k$ steps. The resulting noisy sequence $\tau^{n+k}$ is then fed into both the student model and the teacher model to predict the action sequence $\tau^0$ and $\tau^n$. The target model uses the teacher network’s $k$-step estimation results to predict the action sequence. The student model, trained via consistency distillation and its weights are updated through an Exponential Moving Average (EMA).}
    \label{overview}
\end{figure*}

\section{Related Work}
Our work draws inspiration from a diverse range of topics, including recent advances in diffusion models for robot learning, consistency models, and classic dynamic-system–based methods for robot motion generation.

\textbf{Diffusion Models for Robot Learning:}
\label{II-A}
Recent research in robot learning has increasingly adopted diffusion models as the action decoder to generate smooth and diverse trajectories from noisy inputs. Diffusion models, which were first introduced in computer vision for denoising tasks \cite{ho2020denoising}, have been adapted for robotics by leveraging U-Net based architectures to decode actions from noisy state-action pairs. Early approaches demonstrate that by treating planning as a generative process, the diffusion model can iteratively refine noisy trajectories into feasible control commands \cite{janner2022planning, chi2023diffusion}. More recently, the $\pi_0$ work \cite{black2024pi_0} has further established diffusion models as a robust action decoder, showing that they can effectively map high-dimensional latent representations into precise robotic actions, thereby broadening the applicability of diffusion-based methods in complex control scenarios. Despite its success, existing Diffusion Policy approaches often focus solely on end-to-end trajectory generation without explicitly incorporating structured priors about motion, such as movement primitives. This limits their capability for fast and smooth robot motion generation.

% Diffusion + MPD related works need to rewrite
%The exploration of the most effective architecture for end-to-end robotic manipulation has been a longstanding process. Diffusion Policy frameworks leverage the strengths of diffusion models to learn stochastic policies that capture multimodal action distributions. This approach represents a paradigm shift from deterministic policy representations to probabilistic, noise-driven methodologies capable of encoding diverse action strategies. Additionally, Diffusion Policy has been shown to naturally handle uncertainty, making it suitable for applications in real-world robotics, where noise and variability are prevalent.

%One key advantage of Diffusion Policy is its capacity to generate diverse and feasible trajectories directly from a learned distribution, circumventing the need for extensive reward engineering or handcrafted heuristics. This property has proven particularly useful in tasks like motion planning and manipulation, where the ability to adapt to dynamic environments is crucial.

%Despite its success, existing Diffusion Policy approaches often focus solely on end-to-end trajectory generation without explicitly incorporating structured priors about motion, such as movement primitives. This limits their interpretability and scalability in tasks requiring domain-specific knowledge or hierarchical control. 

\textbf{Consistency Models for Fast Diffusion}
\label{II-B}
Consistency models were originally introduced in the image generation domain to accelerate the sampling speed of diffusion models while maintaining high generation quality~\cite{song2023consistency}. The core principle is to enforce self-consistency along the Probability Flow Ordinary Differential Equation (PF-ODE) trajectory so that any intermediate noisy sample can be directly mapped to the final clean output in a single inference step, as opposed to the multiple denoising steps required in conventional diffusion models. Recent work on Latent Consistency Models (LCMs) has further demonstrated significant speed improvements in text-to-image generation, especially when integrated with diverse conditional control mechanisms~\cite{kim2023latent, dai2024motionlcm, chen2024pixart}. Although consistency models have advanced vision and language synthesis, their use as fast action decoders in robotic motion remains underexplored. To bridge this gap, our work integrates MPs with consistency models to achieve fast, structured, and temporally consistent trajectory generation, thereby extending consistency distillation to robotic motion planning and accelerating robot action inference without compromising success rates.

\textbf{Movement Primitives in Robotics:} \label{II-C}
MPs constitute a fundamental framework in robotics, offering a compact, structured representation for encoding and generating smooth motion trajectories while ensuring temporal coherence. Among the most established formulations, DMPs \cite{ijspeert2013dynamical,saveriano2023dynamic} encode motions as a combination of attractor dynamics and forcing functions, which guarantees spatial and temporal invariance and allows for the generalization of learned trajectories to new goals, yet their reliance on numerical integration limits full trajectory modeling and handling of stochasticity. ProMPs \cite{paraschos2013probabilistic} overcome these issues by representing motions as Gaussian distributions that capture temporal correlations across multiple degrees of freedom, making them effective for learning-from-demonstration. More recently, ProDMPs \cite{prodmps2017} eliminate costly integration via precomputed basis functions, though scaling to high-dimensional motion remains challenging. In contrast to diffusion-based methods that better manage high-dimensional observations, traditional MPs often struggle with scalability in modern robot learning.

\textbf{Diffusion Models using Trajectory Parametrization}
Part of our proposed diffusion framework that applies diffusion to the parametrization weights of trajectories (movement primitives) rather than raw actions or waypoints is also inspired by recent works \cite{carvalho2024motion, scheikl2024movement}, which have shown that coupling diffusion models with movement primitives yields gentle motions for deformable object manipulation and smoother trajectories for general robotic tasks. However, rather than merely combining diffusion models with movement primitives—which still incur inference latency from multi-step denoising—we propose to learn the trajectory-level distribution as the problem to optimize a consistency function that directly maps noise samples to trajectory parameters using consistency distillation. This approach enables fast and smooth action generation.

\section{Methodology}

In this section, we detail our method – Fast Robot Motion Diffusion (FRMD) – and the necessary background. An overview of the Consistency Distillation process is illustrated in Figure \ref{overview}.

\subsection{Problem Formulation}

% Our objective is to directly map observations to structured action sequences, ensuring both efficiency and high quality. Formally, we define the motion trajectory as a sequence of robot actions: $\tau=\{a_i\}_{i=0}^{n}, a_i\in\mathbb{R}^k$ where each action \( a_i \) belongs to a \( k \)-dimensional action space \( \mathbb{R}^k \), determined by control mode and the degrees of freedom (DoFs) of the robot. The goal is to predict the next \( n \) time steps of the trajectory, relative to the current and previous observations. We aim to learn a policy \( \pi \) from an expert demonstration dataset, where the policy maps the robot's observations \( o \in \mathcal{O} \) to an action sequence: $\pi: \mathcal{O} \rightarrow \mathcal{A}$. The observation \( o \) includes RGB images from onboard cameras and robot proprioceptive states, capturing both environmental context and robot dynamics.
Our objective is to directly map observations to structured action sequences while ensuring both efficiency and high quality. Formally, we define the motion trajectory as a sequence of robot actions: $\tau = \{a_i\}_{i=0}^{n},  a_i \in \mathbb{R}^k$
where each action \( a_i \) lies in a \( k \)-dimensional action space \( \mathbb{R}^k \), determined by the control mode and the robot’s degrees of freedom (DoFs). The goal is to predict the next \( n \) time steps of the trajectory based on current and past observations. We seek to learn a policy \( \pi \) from an expert demonstration dataset, mapping the robot’s observations \( o \in \mathcal{O} \) to an action sequence: $\pi: \mathcal{O} \rightarrow \mathcal{A}$. The observation \( o \) comprises RGB images from onboard cameras and robot proprioceptive states, capturing both environmental context and robot dynamics.

\subsection{Preliminaries}

\subsubsection{ Movement Primitives}
\label{III-B}
MPs provide a framework for representing complex motor skills through simple and parameterizable models. ProDMPs offer a unifying framework that overcomes weaknesses and combines the strengths of ProMPs and DMPs. ProDMPs eliminate the need for costly numerical integration associated with DMPs by utilizing precomputed position and velocity basis functions of the fundamental ODE that are valid for all trajectories.

Unlike traditional DMPs, which require solving differential equations for trajectory generation, ProDMPs formulate motion as a weighted combination of precomputed basis functions. In ProDMPs, the positions \( y \) of a trajectory are formulated as:
\begin{equation}
    \begin{aligned}
        y &= c_1 y_1 + c_2 y_2 + 
        \begin{bmatrix} 
            y_2 p_2 - y_1 p_1 & y_2 q_2 - y_1 q_1 
        \end{bmatrix} 
        \begin{bmatrix} 
            w \\ g
        \end{bmatrix} \\
        &= c_1 y_1 + c_2 y_2 + \mathbf{\Phi}^\top \mathbf{w}
    \end{aligned}
\end{equation}
where \( y_1 \) and \( y_2 \) are the two linearly independent complementary functions of the ProDMP’s homogeneous ODE. The constants $c_1$ and $c_2$ are determined by solving a boundary condition problem where we use the current position and velocity, ensuring smooth transitions.

Similar to position $y$, the velocity $\dot{y}$ can be fomulated as:
\begin{equation}
    \dot{y} = c_1 \dot{y_1} + c_2 \dot{y_2} + \dot{\mathbf{\Phi}}^\top \mathbf{w}
\end{equation}
where $\dot{y_1}$, $\dot{y_2}$ are time derivatives version of $y_1$ and $y_2$, respectively. The basis functions for position and velocity, \( \boldsymbol{\Phi} \) and \( \dot{\boldsymbol{\Phi}} \) are predefined and used for motion representation.

The weights \(\mathbf{w}\) are N+1-dim vectors containing the DMP’s original weight vector with the goal attractor to which the ODE converges. ProDMPs facilitate planning smooth trajectories with guaranteed boundary conditions while minimizing computational demands.

\subsubsection{Consistency Models}
\label{III-C}
The Consistency Model introduces an efficient generative model designed for effective single-step or few-step inference generation while maintaining a comparable performance. Consistency models are built upon the PF-ODE, which describes the evolution of data distribution over time. Given an data $x_t$ at time $t$, the PF-ODE is defined as:
\begin{equation}
    dx = -\dot{\sigma}(t) \sigma(t) \nabla_{x} \log p(x_t) dt
    \label{ODE}
\end{equation}
where \( x_t \) is the noisy observation at time \( t \), \( \sigma(t) \) is the noise schedule and \( \nabla_{x_t} \log p(x_t) \) is the score function, guiding the denoising process.

The objective of consistency model is to learn the solution function \( f(\cdot, \cdot) \) of this PF-ODE. Given a solution trajectory \( \{\mathbf{x}_t\}_{t \in [0, T]} \) of the PF-ODE, the consistency function \( f \) is defined as \( f : (\mathbf{x}_t, t) \mapsto \mathbf{x}_0 \) that directly maps any noisy input $x_t$ at time $t$ to the original clean data $x_0$, enforcing the self-consistency property:
\begin{equation}
    f(\mathbf{x}_t, t) = f(\mathbf{x}_{t'}, t'), \quad \forall t, t' \in [\epsilon, T]
\label{1}
\end{equation}

As shown in Equation \ref{1}, the implication of the self-consistency property is that for any input pairs \( (\mathbf{x}_t, t) \) on the same PF-ODE trajectory, their outputs \( f(\mathbf{x}_t, t) \) remain consistent. All consistency models have to meet the boundary condition \( f(\mathbf{x}_0, t_0) = \mathbf{x}_0 \). The boundary condition ensures that the model does not converge to a meaningless solution like \( f_{\theta}(\mathbf{x}, t) \equiv 0 \).

\textit{Consistency Distillation} is a method widely used for training consistency model by distilling knowledge from pre-trained diffusion models (teacher model). The consistency loss is defined as:
\begin{equation}
    \label{eq.5}
    \mathcal{L}(\theta, \theta^{-}; \mathbf{\phi}) = \mathbb{E} \left[ d \left( f_{\theta}(\mathbf{x}_{t_{n+1}}, t_{n+1}), f_{\theta^{-}}(\hat{\mathbf{x}}_{t_n}^{\mathbf{\phi}}, t_n) \right) \right]
\end{equation}
where \( d(\cdot, \cdot) \) is a metric function chosen for measuring the distance between two samples. \( \phi(\cdot, \cdot) \) is the update function of ODE solver applied to the PF-ODE. \( f_{\theta}(\cdot, \cdot) \) and \( f_{\theta^{-}}(\cdot, \cdot) \) are referred to as ‘online network’ and ‘target network’ according to . When the optimization of the online network converges, the target network will eventually match the online network since \( \theta^{-} \) is a running average of \( \theta \). The estimated consistency model can become arbitrarily accurate as long as the step size of the ODE solver is sufficiently small and the consistency distillation loss reaches zero.

\subsection{FRMD: Fast Robot Motion Diffusion}

The overvew of FRMD is presented in Figure \ref{overview}. In the pre-training phase, we train the teacher model follow the pipeline proposed in \cite{scheikl2024movement}. Then FRMD implements a consistency distillation method to distill the knowledge from the teacher diffusion-based policy. We adopt a consistency function to predict the action sample, opposite to the noise prediction that is commonly employed in image generation. This modification results in a faster convergence to the low-dimensional robot action manifold. In the inference phase, FRMD is able to decode high-quality action within one inference.

\subsubsection{Teacher Model Set-up} For Teacher Model, we follow the pipeline proposed in \cite{scheikl2024movement}. The teacher model consists of a trainable model $E_\theta$ that outputs a weight vector $w$. Combined with initial values $y_0$, $\dot{y_0}$ for position and velocity, $w$ is decoded into an action sequence $\tau$ using ProDMP decoder $P$ with predefined parameters $\boldsymbol{\Phi}$ as mentioned in \ref{III-B}. Assume that the noise action is $\tilde{\tau}$, the teacher model denoise pipeline can be represented as:
\begin{equation}
\label{eq.6}
     F_\theta(\tilde{\tau}, o, t) = P_\Phi(y_0, \dot{y_0}, E_\theta(\tilde{\tau}, o, t))
\end{equation}

For the diffusion part, we adopt score-based generative model which is also built upon PF-ODE, here we rewrite the PF-ODE in Equation \ref{ODE} as follows:

\begin{equation}
    \label{eq.7}
    d\tau= -\dot{\sigma}(t) \sigma(t) \nabla_{\tau} \log p(\tau | o, \sigma(t)) dt
\end{equation}
where $\sigma(t)$ represents the noise scheduler and the score function $\nabla_{\tau} \log p(\tau | o, \sigma(t))$ can be seen as the gradient of the probability of action sequences $\tau$ conditioned by observations $o$ and $\sigma(t)$. According to Equation. \ref{eq.7}, the score function can be approximated as:
\begin{equation}
    \nabla_{\tau} \log p(\tau | o, \sigma(t))  \doteq \frac{F_{\theta}(\tau, o, \sigma(t)) - \tau}{\sigma(t)^2}
\end{equation}

During training, we adopt score matching method to minimize the loss function: 
\begin{equation}
    \mathbb{E}\left[ \left\| \frac{F_{\theta}(\tilde{\tau}, o, \sigma(t)) - \tilde{\tau}}{\sigma(t)^2} - \nabla_{\tilde{\tau}} \log q(\tilde{\tau} | \tau) \right\|^2 \right]
\end{equation}
where $\tilde{\tau}=\tau+\epsilon$ with Gaussian noise \( \epsilon \sim \mathcal{N}(0, t^2 I) \).

During inference, new actions are generated by gradually denoising samples of a unit Gaussian by solving the PF-ODE using DPM-Solver \cite{lu2022dpm} that are specifically designed for fast inference($\sim$10 steps) in ODE-based diffusion.

\subsubsection{Student Model Consistency Distillation}

To train student model, we adopt the method of consistency distillation, as depicted in Figure \ref{overview}. Given a ground-truth expert action sequence \( \tau_0 \), we first obtain the noisy action \( \tau^{n+k} \) by conducting a forward operation with \( n + k \) steps to add noise on \( \tau^0 \), where \( k \) is the skipping interval in sampling of teacher model. Subsequently, the noisy action \( \tau^{n+k} \) is feed-forwarded to teacher model and student model, respectively. The student model directly predict clean action sequence $\hat{\tau^0}$. The teacher model output action sequence $\hat{\tau}^{n}$ via $k$-step estimation:
\begin{equation}
    \hat{\tau^{n}} \leftarrow \tau^{n+k} + (t_n - t_{n+k}) \phi (\tau^{n+k}, t_{n+k})
    \label{2}
\end{equation}

In order to make sure the self-consistency property (Equation \ref{1}) holds, we also design a target model cloned from the student model, and we expect the outputs of the student model and the target model to be aligned. Specifically, the target network generates \( \hat{\tau^{0-}} \) utilizing \( \hat{\tau^n} \) and we expect that \( \hat{\tau^{0-}}=\hat{\tau^0} \).

In the student and target model, clean action sequence \( \hat{\tau}^{0-}\) and \(\hat{\tau}^0 \) are obtained by consistency function. We parameterize the consistency function using skip connections as mentioned in \cite{song2023consistency}:
\begin{equation}
    \label{eq.11}
    f_{\theta}(\mathbf{\tau}, o, t) = c_{\text{skip}}(t) \mathbf{\tau} + c_{\text{out}}(t) F_{\theta}(\mathbf{\tau}, o, t)
\end{equation}
where the $F_\theta(\cdot, \cdot, \cdot)$ is the same structure as teacher model proposed in Equation. \ref{eq.6}. Meanwhile, to strength the boundary condition mentioned in \ref{III-C}, we set $c_{\text{skip}}(t) = {\gamma_d^2}/(\beta^2 t^2 + \gamma_d^2)$ and $c_{\text{out}}(t) = {\beta t}/{\sqrt{\beta^2 t^2 + \gamma_d^2}}$. where $\beta$ denotes scaling value while $\gamma_d$ is a balance value. Combined with Equation. \ref{eq.5} and Equation. \ref{eq.11}, the consistency distillation loss can be expressed as:
\begin{equation}
\begin{aligned}
    \mathcal{L}_{CD} = \\
    &\mathbb{E} \big[ \lambda(t_n) d(f_{\theta}(\mathbf{\tau}^{n+k}, o, t_{n+k}), f_{\theta^-} (\hat{\mathbf{\tau}}^n, o, t_n)) \big]
\end{aligned}
\end{equation}
where \( \lambda(\cdot) \in \mathbb{R}^+ \) is a positive weighting function, \( \hat{\mathbf{\tau}}^{n} \) is given by Equation \ref{2}, \( d(\cdot, \cdot) \) is a metric function that satisfies $\forall \mathbf{x}, \mathbf{y}:d(\mathbf{x}, \mathbf{y}) \geq 0 \text{ and } d(\mathbf{x}, \mathbf{y})=0 \text{ if and only if } \mathbf{x} = \mathbf{y}.$  \( \theta^- \) is updated with the exponential moving average (EMA) of the parameter \( \theta \), defined as:
\begin{equation}
    \theta^- \leftarrow \text{stopgrad}(\mu \theta^- + (1 - \mu) \theta),
\end{equation}
where \( \text{stopgrad}(\cdot) \) denotes the stop-gradient operation and \( \mu \) satisfies \( 0 \leq \mu < 1 \).

To summarize, we propose Student Model Consistency Distillation as detailed in Algorithm\ref{alg1}.

\begin{algorithm}[htbp]
	%\textsl{}\setstretch{1.8}
	% \renewcommand{\algorithmicrequire}{\textbf{Input:}}
	% \renewcommand{\algorithmicensure}{\textbf{Output:}}
	\caption{Student Model Consistency Distillation}
	\label{alg1}
	\begin{algorithmic}[1]
		\STATE Initialization: Dataset $ \mathcal{D} $, initial parameter $ \theta $, learning rate $ \eta $, ODE solver $ \phi(\cdot, \cdot) $, $ d(\cdot, \cdot) $, $ \lambda(\cdot) $ and $ \mu $
		\REPEAT
		\STATE Sample $ o, \mathbf{\tau^0} \sim \mathcal{D} $ and $ n \sim \mathcal{U}\left[1, N - k\right] $
            \STATE Sample $ \mathbf{\tau}^{n+k} \sim \mathcal{N}(\mathbf{\tau}; t_{n+k}^2 \mathbf{I}) $
            \STATE Teacher Model $k$-step Denoise:\\
            $\hat{\tau}^{n} \leftarrow \tau^{n+k} + (t_n - t_{n+k}) \phi (\tau^{n+k}, t_{n+k})$
            \STATE Compute loss:
            \begin{equation*}
                \mathcal{L}_{CD} \gets \lambda(t_n) d(f_{\theta}(\mathbf{\tau}^{n+k}, o, t_{n+k}), f_{\theta^-} (\hat{\mathbf{\tau}}^{n}, o, t_n))
            \end{equation*}
            \STATE Update student model:\\
            $ \theta \gets \theta - \eta \nabla_{\theta} \mathcal{L}_{CD}(\theta, \theta^-; \boldsymbol{\phi}) $
            \STATE Update target model using EMA:\\
            $\theta^- \gets \text{stopgrad}(\mu \theta^- + (1 - \mu) \theta)$ 
            
            \UNTIL convergence
	\end{algorithmic}  
\end{algorithm}

\section{Evaluations}
% evaluation objectives by answering three questions
\label{IV}
In the evaluation part, we aim to answer the following questions: 

(1) Can our proposed FRMD generate smooth and temporally consistent trajectories across different environments and various maniplation tasks ranging from simple to complex? 

(2) Can our proposed FRMD achieve one-step inference without sacrificing motion quality in these benchmark tasks? 

(3) How much additional performance gain can we achieve when comparing FRMD with other SOTA baseline methods? 

(4) How will different backbone network architecture affect the final performance of our proposed FRMD?
\begin{figure*}[htbp]
    \centering
    \includegraphics[width=1\textwidth]{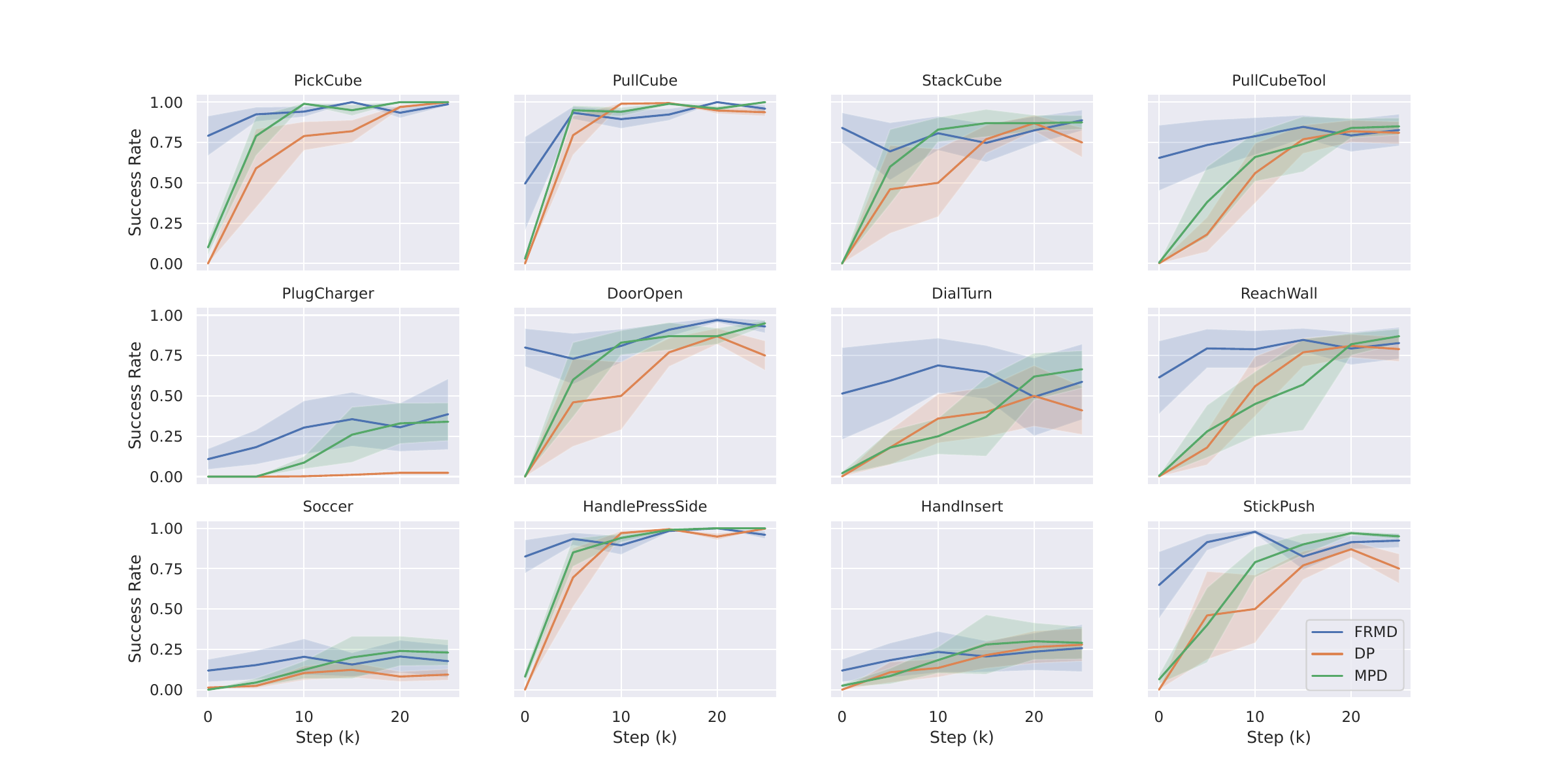}
    \caption{\textbf{Learning Curve comparison of different methods across various robotic tasks.} We compare FRMD (ours), Diffusion Policy (DP) and Movement Primitives Diffusion (MPD) across 12 different tasks from the MetaWorld and ManiSkill benchmarks. The success rate is computed by evaluating 10 episodes with random seeds in each environment at every 5k training steps for each method, until convergence. The mean success rate is plotted as a solid line, and the variance is shown as shaded areas. Results show that our method consistently achieves higher success rates with significantly smaller inference latency (10x faster than MPD~\cite{scheikl2024movement}, 7x faster than DP~\cite{chi2023diffusion}, as shown in Table 1) compared to baselines. Note that the intermediate success rate of our method in the initial training steps is due to model distillation using the teach model.}
    \label{success rate}
\end{figure*}
\subsection{Environments, Tasks and Datasets}

We conduct our experiments in the well-recognized MetaWorld~\cite{yu2020meta} and Maniskills~\cite{mu2021maniskill} benchmarks. Following ~\cite{lu2024manicm}, we use a total of 12 tasks ranging from simple tasks like pick-cube to more challenging ones such as dexterous manipulation. Like ~\cite{lu2024manicm}, we divided the 12 tasks into three categories, including 4 easy tasks, 5 medium tasks and 3 hard tasks based on their difficulty levels. For datasets collection, we collect demonstrations by running expert policies in Metaworld and replaying trajectory method in Maniskills. Each demonstration is a sequence pair $(\tau_i, o_i)$ over one full task execution with $N$ time steps. During preprocessing, the demonstration datasets are split into multiple action and observation sequences of lengths $n$ and $m$, respectively. So that the model will predict next $n$-step actions based on observations from the previous $m$ time steps. We obtain a number of 100 expert demonstrations for training in each benchmark for fair comparisons.

\subsection{Evaluation Metrics}
\label{IV-B}
To answer the (1)-(3) questions proposed in the beginning of section \ref{IV}, we report three key metrics to assess the performance of our method: (1) Task Metrics, which measure the success rate on benchmark tasks, (2) Time Metrics, which evaluate the efficiency of inference, and (3) Motion Smoothness Metrics, which assess the quality and smoothness of generated motions.

\subsubsection{Task Metrics} We evaluate 10 episodes every 5,000 training steps and compute the average success rates(SR) throughout the training process and until each training converges. For success rate, the higher is the better.

\subsubsection{Time Metrics}In the time assessment phase, we measure the average runtime per step. To mitigate performance fluctuations, each main experiment is conducted using three different random seeds (0, 1, and 2). For inference time, lower values indicate better performance.

\subsubsection{Motion Quality Metrics} Defining Motion Quality Metrics is challenging since directly measuring the model's inference consistency involves
complex statistical methods\cite{jin2020geometric}\cite{flury1996self}. 
We adopt a geometric approach~\cite{guillen2020measuring} to visualize generated trajectories and assess their smoothness by computing the curvature $k_i$  at each data point. A transition is classified as non-smooth if $k_i > k_{max} = 1$. Finally, we visualize these non-smooth transitions in our visualizations and compare our method with baselines.

\subsection{Baselines}Our work primarily focuses on two key contributions: accelerating inference speed and improving the quality of robotic action generation. To evaluate our approach, we compare it against state-of-the-art baselines, including DP and MPD. Notably, MPD serves as the teacher model in our consistency distillation framework. We make comparison between our method and teacher model to see if our model could surpasses MPD by achieving significantly faster inference while maintaining or even enhancing action quality.

\subsection{Implementation Details}
\subsubsection{Dataset} We predict action sequences of length \( n = 12 \), conditioned on the previous \( m = 3 \) observations. As the result, the demonstrations are split into multiple action and observation sequences of lengths 12 and 3, respectively.
\subsubsection{Observation Encoder}
The encoder maps the raw image sequence into a latent embedding $o_t$ and is trained end-to-end with our method and baselines. We used a standard ResNet18 (without pretraining and with output size 128) as the encoder. As decribed in \cite{chi2023diffusion}\cite{chi2023diffusionpolicy}, we also made some modifications: Replace the global average pooling with a spatial softmax pooling and replace BatchNorm with GroupNorm.

\subsubsection{Network Architectures}
For Diffusion Policy, we use CNN layer with sizes of (256, 512, 1024). MPD and FRMD share an optimal transformer architecture, as proposed in \cite{chi2023diffusion}, with 6 layers, 4 heads, dropout probability of 0.3 and embedding size of 256.

\subsubsection{Training}
An AdamW optimizer with a batch size of 128, a learning rate of 1e-4 and a weight decay of 1e-6 is employed for 30k steps training. We also adopt a cosine decay learning rate scheduler and 500 iterations of linear warm-up. The EMA rate is set to $\mu$ = 0.95. We implement our model in PyTorch, and train the model on one NVIDIA RTX 4090 GPU. All compared models are evaluated on the same device to achieve fairness.

\begin{table}[htbp]
    \centering
    \small
    \caption{Success rate and inference time across all tasks based on our models and baselines. The best results for each category are in bold font and the second best ones are underlined for an easier comparison.}
    \label{tab}
    \renewcommand{\arraystretch}{1.4}
    \begin{tabular}{lccccc}
        \toprule
        \textbf{Methods} & \textbf{Easy(4)} & \textbf{Medium(5)} & \textbf{Hard(3)} & \textbf{Average} \\
        \midrule
        \multicolumn{5}{c}{\textbf{Success Rate (\%)}} \\
        DP  & \textbf{99.3}$\pm${0.1}  & 41.0$\pm${3.2}  & 10.1$\pm${1.4}  & 50.1 \\
        MPD   & 98.9$\pm${0.3} & \underline{64.8}$\pm${2.6} & \underline{28.6}$\pm${2.9} &  \underline{64.1} \\
        FRMD & \underline{99.2}$\pm${0.1} & \textbf{66.3}$\pm${1.2} & \textbf{29.0}$\pm${2.3} & \textbf{64.8}  \\
        
        \midrule
        \multicolumn{5}{c}{\textbf{Inference Time (ms)}} \\
        DP  & \underline{119.8}$\pm${1.4} & \underline{121.3}$\pm${2.3}  & \underline{118.2}$\pm${3.6} & \underline{119.7} \\
        MPD  & 162.7$\pm${3.5} & 173.2$\pm${3.6} & 169.9$\pm${1.2}  & 168.6 \\
        FRMD  & \textbf{15.2}$\pm${0.8} & \textbf{18.6}$\pm${3.4}  & \textbf{17.9}$\pm${1.1} & \textbf{17.2} \\
        
        \bottomrule
    \end{tabular}
\end{table}

\subsection{Results and Comparisons}
We compare FRMD with existing state-of-the-art methods on Maniskills and Metaworld to demonstrate the performence of our method. As described in \ref{IV-B}, we make the comparison on three parts: Success Rate, Inference time and Motion quality. According to these result, we have the following analysis concerning the three questions proposed in the start of Section \ref{IV}.

\subsubsection{Success Rate} Figure. \ref{success rate} presents the learning curve of our method and baselines. The results show that our method accelerates the diffusion process without any performance drop. Specifically, with only one-step inference, our FRMD can approximate or even surpass the state-of-the-art model. The detailed results are presented on the Table \ref{tab}, that FRMD achieves the highest overall success rate of 64.8\%, outperforming both MPD (64.1\%) and DP (50.1\%). For easy tasks, FRMD (99.2\%) performs on par with DP (99.3\%) and slightly better than MPD (98.9\%), indicating that all methods perform well when task complexity is low. As the task difficulty increases, FRMD consistently maintains its advantage, achieving 66.3\% success on medium tasks, confirming that our method effectively distills MPD’s structured motion representation into a more efficient form. On hard tasks, FRMD continues to outperform MPD and significantly surpasses DP, demonstrating its superior ability to handle complex robotic motion generation. These results provide strong evidence that while FRMD is built upon MPD as its teacher model, it ultimately surpasses MPD in performance, successfully retaining the structured motion quality while enhancing policy expressiveness and robustness.

\subsubsection{Inference Time} The inference time results highlight the efficiency advantage of FRMD, which achieves real-time action generation with an average inference time of 17.2ms, making it 10× faster than MPD (168.6ms) and 7× faster than DP (119.7ms). Across different task complexities, FRMD maintains consistently low latency, achieving 15.2ms on easy tasks, 18.6ms on medium tasks, and 17.9ms on hard tasks, demonstrating its ability to efficiently generate high-quality actions in a single step. These results confirm that FRMD’s consistency-distilled model significantly reduces computational overhead while preserving high motion quality, making it well-suited for real-time robotic control applications.

\subsubsection{Motion Quality} 
\begin{figure}[htbp]
    \centering
    \includegraphics[width=0.49\textwidth]{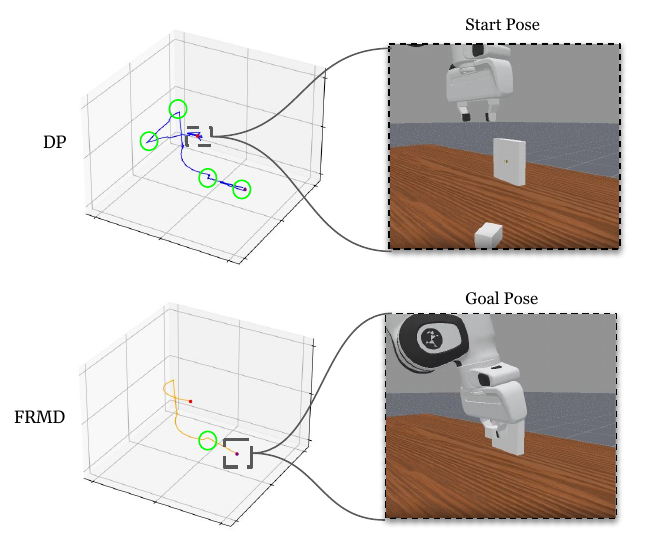}
    \caption{\textbf{Trajectories generated in PlugCharger-v1 task.} The top plot shows the trajectory generated by DP, while the bottom plot presents the trajectory produced by our method (FRMD). The green circles highlight regions where the data point transitions exhibits significant non-smoothness which is computed by comparing its curvature $k_i$ with a threshold $k_{max}=1$, as defined in Section \ref{IV-B} . In comparison, our method results in a significantly smoother trajectory with fewer oscillations, demonstrating improved motion stability, especially near the start and goal point.}
    \label{trajectory}
\end{figure}

To further evaluate the motion quality of different methods, we visualize the end-effector trajectory during execution. To ensure fairness, we fix the environment's initial conditions before conducting experiments, maintaining the same initial and goal states. The visualization results for the PlugCharger-v1 task are shown in Figure~\ref{trajectory}. As described in Section~\ref{IV-B}, a transition is classified as non-smooth if the curvature satisfies \( k_i > 1 \). We record the number of non-smooth transitions as \( N = num \{ k_i > 1 \} \) in the generated trajectory. The results show that \( N_{\text{DP}} = 82 \) and \( N_{\text{FRMD}} = 21 \). Given that the maximum episode step is fixed across all experiments, this indicates that our method (FRMD) produces significantly smoother trajectories compared to DP.

Based on the above results analysis, accordingly, we conclude the answers to Question(1)-(3) raised before: 

(1) Our proposed FRMD effectively generates smooth and temporally consistent trajectories across different environments, significantly reducing non-smooth transitions compared to DP. 

(2) Our FRMD achieves one-step inference with an average latency of 17.2ms, which is 10× faster than MPD, without compromising motion quality. 

(3) Compared to SOTA baselines, our FRMD achieves the highest success rate and demonstrates superior efficiency and performance.

\subsection{Ablation Study}
We also conducted ablation studies to further evaluate the impact of different design choices in our method. Specifically, we aim to answer Question (4) raised at the beginning of this section: \textit{How will different backbone network architecture affect the final performance of our proposed FRMD?}

We investigate the impact of different backbone architectures on the performance of FRMD. Our approach adopts a transformer-based model~\cite{dasari2024ingredients} as the backbone for the teacher model (MPD~\cite{scheikl2024movement}). Specifically, the denoiser network is implemented as a transformer with 6 layers, 4 attention heads, and a hidden size of 256, enabling it to effectively capture temporal dependencies in robotic trajectories. To evaluate the effectiveness of this architecture, we compare it against two alternative designs: (1) a lightweight transformer with fewer parameters and (2) a simple feedforward MLP network composed of fully connected layers. To ensure a fair comparison, all other components (e.g., observation encoders) remain unchanged across experiments.

The comparison results are presented in Table \ref{ablation1}. The original version of the Transformer achieves the highest success rates but at the cost of slightly higher inference time. The lightweight Transformer (Transformer* in Table \ref{ablation1}) achieves lower inference time while maintaining strong performance across tasks. MLP is computationally efficient but significantly underperforms in complex tasks. These results validate that Transformer-based architectures are well-suited for our method, enabling the generation of fast and temporally consistent robot motions.

\begin{table}[htbp]
    \centering
    % \scriptsize
    \small
    \caption{Ablation study on different architectures. Here Transformer* refers to a lightweight transformer with the same architecture as the Transformer baseline but with fewer parameters. The best results are highlighted in bold.}
    \label{tab:teacher_ablation}
    \renewcommand{\arraystretch}{1.4}
    \begin{tabular}{l|cc|cc|cc}
        \toprule
        \textbf{Architecture} & \multicolumn{2}{c|}{\textbf{Easy(4)}} & \multicolumn{2}{c|}{\textbf{Medium(5)}} & \multicolumn{2}{c}{\textbf{Hard(3)}} \\
        & \textbf{Time} & \textbf{SR} & \textbf{Time} & \textbf{SR} & \textbf{Time} & \textbf{SR} \\
        \midrule
        Transformer& 15.2 & \textbf{99.2} & 18.6 & 66.3 & 17.9 & \textbf{29.0} \\
        \midrule
        Transformer*& 16.2 & 98.5 & 15.8 & \textbf{68.5} & 15.9 & 26.2 \\
        \midrule
        MLP & \textbf{9.9} & 74.9 & \textbf{10.3} & 54.2 & \textbf{8.9} & 9.2 \\
        \bottomrule
    \end{tabular}
    \label{ablation1}
\end{table}

% \subsubsection{Score function}
%     For the score function. An alternative is diffusing action sequences in the ProDMP’s weight space to approximate the score function of weight vector distribution $p(w | o, \sigma(t))$ instead of action sequence distribution $p(\tau | o, \sigma(t))$. In preliminary experiments, we found that diffusing in weight space does not reach the high success rates of diffusing action sequences directly. Diffusing action sequences can leverage sophisticated network architectures such as transformers that benefit from the sequential nature of action sequences, which are not explicit in weight space.

\section{Conclusions and Limitations}
In this work, we propose FRMD, a novel consistency-distilled movement primitives model that integrates Consistency Models with Probabilistic Dynamic Movement Primitives (ProDMPs~\cite{paraschos2013probabilistic}) for efficient, smooth robotic motion generation. Using Movement Primitive Diffusion (MPD~\cite{scheikl2024movement}) as the teacher model, our approach learns to produce structured motion trajectories while dramatically improving inference speed via consistency distillation. Unlike conventional diffusion policies that require multiple denoising steps, FRMD achieves real-time, single-step inference without sacrificing trajectory quality. In future work, we plan to extend FRMD to complex, high-dimensional tasks and integrate task-specific cost functions into the consistency training framework to further evaluate its applicability in more robotic tasks.

Due to time limits, we did not test our method on large-scale task pre-training using large datasets and large parameterized networks. It's worth noting that our primary contribution --- the design of the action decoder --- is inherently modular and can be seamlessly integrated into large-scale diffusion-based VLA (vision-language-action) models~\cite{dasari2024ingredients, black2024pi_0}. By employing our method as the action decoder (``action expert'') and pairing it with a powerful encoders, such as the one used in the $\pi_0$~\cite{black2024pi_0} model, our framework holds significant potential for improved performance and scalability across diverse robotic tasks.


%%%%%%%%%%%%%%%%%%%%%%%%%%%%%%%%%%%%%%%%%%%%%%%%%%%%%%%%%%%%%%%%%%%%%%%%%%%%%%%%

%%%%%%%%%%%%%%%%%%%%%%%%%%%%%%%%%%%%%%%%%%%%%%%%%%%%%%%%%%%%%%%%%%%%%%%%%%%%%%%%

%%%%%%%%%%%%%%%%%%%%%%%%%%%%%%%%%%%%%%%%%%%%%%%%%%%%%%%%%%%%%%%%%%%%%%%%%%%%%%%%

% \section{Acknowledgment}

% The authors would like to acknowledge Yi Hu and Yafei Ou from University of Alberta for their helpful discussion and suggestions.

% \printbibliography
% \bibliographystyle{ieee}
\bibliography{references}

\begin{thebibliography}{10}
\providecommand{\url}[1]{#1}
\csname url@samestyle\endcsname
\providecommand{\newblock}{\relax}
\providecommand{\bibinfo}[2]{#2}
\providecommand{\BIBentrySTDinterwordspacing}{\spaceskip=0pt\relax}
\providecommand{\BIBentryALTinterwordstretchfactor}{4}
\providecommand{\BIBentryALTinterwordspacing}{\spaceskip=\fontdimen2\font plus
\BIBentryALTinterwordstretchfactor\fontdimen3\font minus \fontdimen4\font\relax}
\providecommand{\BIBforeignlanguage}[2]{{%
\expandafter\ifx\csname l@#1\endcsname\relax
\typeout{** WARNING: IEEEtran.bst: No hyphenation pattern has been}%
\typeout{** loaded for the language `#1'. Using the pattern for}%
\typeout{** the default language instead.}%
\else
\language=\csname l@#1\endcsname
\fi
#2}}
\providecommand{\BIBdecl}{\relax}
\BIBdecl

\bibitem{chi2023diffusion}
C.~Chi, Z.~Xu, S.~Feng, E.~Cousineau, Y.~Du, B.~Burchfiel, R.~Tedrake, and S.~Song, ``Diffusion policy: Visuomotor policy learning via action diffusion,'' \emph{The International Journal of Robotics Research}, p. 02783649241273668, 2023.

\bibitem{dasari2024ingredients}
S.~Dasari, O.~Mees, S.~Zhao, M.~K. Srirama, and S.~Levine, ``The ingredients for robotic diffusion transformers,'' \emph{arXiv preprint arXiv:2410.10088}, 2024.

\bibitem{dong2023aligndiff}
Z.~Dong, Y.~Yuan, J.~Hao, F.~Ni, Y.~Mu, Y.~Zheng, Y.~Hu, T.~Lv, C.~Fan, and Z.~Hu, ``Aligndiff: Aligning diverse human preferences via behavior-customisable diffusion model,'' \emph{arXiv preprint arXiv:2310.02054}, 2023.

\bibitem{black2024pi_0}
K.~Black, N.~Brown, D.~Driess, A.~Esmail, M.~Equi, C.~Finn, N.~Fusai, L.~Groom, K.~Hausman, B.~Ichter \emph{et~al.}, ``$\pi_0 $: A vision-language-action flow model for general robot control,'' \emph{arXiv preprint arXiv:2410.24164}, 2024.

\bibitem{carvalho2024motion}
J.~Carvalho, A.~Le, P.~Kicki, D.~Koert, and J.~Peters, ``Motion planning diffusion: Learning and adapting robot motion planning with diffusion models,'' \emph{arXiv preprint arXiv:2412.19948}, 2024.

\bibitem{ho2020denoising}
J.~Ho, A.~Jain, and P.~Abbeel, ``Denoising diffusion probabilistic models,'' in \emph{Advances in Neural Information Processing Systems}, 2020.

\bibitem{song2020denoising}
J.~Song, C.~Meng, and S.~Ermon, ``Denoising diffusion implicit models,'' \emph{arXiv preprint arXiv:2010.02502}, 2020.

\bibitem{li2023prodmp}
G.~Li, Z.~Jin, M.~Volpp, F.~Otto, R.~Lioutikov, and G.~Neumann, ``Prodmp: A unified perspective on dynamic and probabilistic movement primitives,'' \emph{IEEE Robotics and Automation Letters}, vol.~8, no.~4, pp. 2325--2332, 2023.

\bibitem{paraschos2013probabilistic}
A.~Paraschos, C.~Daniel, J.~Peters, and G.~Neumann, ``Probabilistic movement primitives,'' in \emph{Proceedings of the IEEE International Conference on Robotics and Automation (ICRA)}, 2013, pp. 115--122.

\bibitem{scheikl2024movement}
P.~M. Scheikl, N.~Schreiber, C.~Haas, N.~Freymuth, G.~Neumann, R.~Lioutikov, and F.~Mathis-Ullrich, ``Movement primitive diffusion: Learning gentle robotic manipulation of deformable objects,'' \emph{IEEE Robotics and Automation Letters}, 2024.

\bibitem{song2023consistency}
Y.~Song, P.~Dhariwal, M.~Chen, and I.~Sutskever, ``Consistency models,'' 2023.

\bibitem{song2023improved}
Y.~Song and P.~Dhariwal, ``Improved techniques for training consistency models,'' \emph{arXiv preprint arXiv:2310.14189}, 2023.

\bibitem{schaal2005learning}
S.~Schaal, J.~Peters, J.~Nakanishi, and A.~Ijspeert, ``Learning movement primitives,'' in \emph{Robotics Research. The Eleventh International Symposium: With 303 Figures}.\hskip 1em plus 0.5em minus 0.4em\relax Springer, 2005, pp. 561--572.

\bibitem{saveriano2023dynamic}
M.~Saveriano, F.~J. Abu-Dakka, A.~Kramberger, and L.~Peternel, ``Dynamic movement primitives in robotics: A tutorial survey,'' \emph{The International Journal of Robotics Research}, vol.~42, no.~13, pp. 1133--1184, 2023.

\bibitem{yu2020meta}
T.~Yu, D.~Quillen, Z.~He, R.~Julian, K.~Hausman, C.~Finn, and S.~Levine, ``Meta-world: A benchmark and evaluation for multi-task and meta reinforcement learning,'' in \emph{Conference on robot learning}.\hskip 1em plus 0.5em minus 0.4em\relax PMLR, 2020, pp. 1094--1100.

\bibitem{mu2021maniskill}
T.~Mu, Z.~Ling, F.~Xiang, D.~Yang, X.~Li, S.~Tao, Z.~Huang, Z.~Jia, and H.~Su, ``Maniskill: Generalizable manipulation skill benchmark with large-scale demonstrations,'' \emph{arXiv preprint arXiv:2107.14483}, 2021.

\bibitem{lu2024manicm}
G.~Lu, Z.~Gao, T.~Chen, W.~Dai, Z.~Wang, and Y.~Tang, ``Manicm: Real-time 3d diffusion policy via consistency model for robotic manipulation,'' \emph{arXiv preprint arXiv:2406.01586}, 2024.

\bibitem{janner2022planning}
M.~Janner, Q.~Yang, and S.~Levine, ``Planning with diffusion for flexible behavior synthesis,'' in \emph{International Conference on Learning Representations}, 2022.

\bibitem{kim2023latent}
J.~Kim, K.~Lee, and J.~Park, ``Latent consistency models for accelerated text-to-image generation,'' \emph{arXiv preprint arXiv:2301.00000}, 2023.

\bibitem{dai2024motionlcm}
W.~Dai, L.-H. Chen, J.~Wang, J.~Liu, B.~Dai, and Y.~Tang, ``Motionlcm: Real-time controllable motion generation via latent consistency model,'' in \emph{European Conference on Computer Vision}.\hskip 1em plus 0.5em minus 0.4em\relax Springer, 2024, pp. 390--408.

\bibitem{chen2024pixart}
J.~Chen, Y.~Wu, S.~Luo, E.~Xie, S.~Paul, P.~Luo, H.~Zhao, and Z.~Li, ``Pixart-$\{$$\backslash$delta$\}$: Fast and controllable image generation with latent consistency models,'' \emph{arXiv preprint arXiv:2401.05252}, 2024.

\bibitem{ijspeert2013dynamical}
A.~J. Ijspeert, J.~Nakanishi, H.~Hoffmann, P.~Pastor, and S.~Schaal, ``Dynamical movement primitives: learning attractor models for motor behaviors,'' \emph{Neural computation}, vol.~25, no.~2, pp. 328--373, 2013.

\bibitem{prodmps2017}
J.~Doe and J.~Smith, ``Probabilistic dynamic movement primitives for efficient robotic trajectory generation,'' in \emph{Proceedings of the IEEE International Conference on Robotics and Automation (ICRA)}, 2017, pp. 4567--4573.

\bibitem{lu2022dpm}
C.~Lu, Y.~Zhou, F.~Bao, J.~Chen, C.~Li, and J.~Zhu, ``Dpm-solver: A fast ode solver for diffusion probabilistic model sampling in around 10 steps,'' \emph{Advances in Neural Information Processing Systems}, vol.~35, pp. 5775--5787, 2022.

\bibitem{jin2020geometric}
J.~Jin, L.~Petrich, M.~Dehghan, and M.~Jagersand, ``A geometric perspective on visual imitation learning,'' in \emph{2020 IEEE/RSJ International Conference on Intelligent Robots and Systems (IROS)}.\hskip 1em plus 0.5em minus 0.4em\relax IEEE, 2020, pp. 5194--5200.

\bibitem{flury1996self}
B.~Flury and T.~Tarpey, ``Self-consistency: A fundamental concept in statistics,'' \emph{Statistical Science}, vol.~11, no.~3, pp. 229--243, 1996.

\bibitem{guillen2020measuring}
S.~Guill{\'e}n~Ruiz, L.~V. Calderita, A.~Hidalgo-Paniagua, and J.~P. Bandera~Rubio, ``Measuring smoothness as a factor for efficient and socially accepted robot motion,'' \emph{Sensors}, vol.~20, no.~23, p. 6822, 2020.

\bibitem{chi2023diffusionpolicy}
C.~Chi, S.~Feng, Y.~Du, Z.~Xu, E.~Cousineau, B.~Burchfiel, and S.~Song, ``Diffusion policy: Visuomotor policy learning via action diffusion,'' in \emph{Proceedings of Robotics: Science and Systems (RSS)}, 2023.

\end{thebibliography}
\end{document}